\documentclass[letterpaper, 10 pt, conference]{ieeeconf}  
\usepackage[style=ieee]{biblatex}
\usepackage{amsmath}
\usepackage{amssymb}
\usepackage{graphicx}
\usepackage{multirow} 
\usepackage{hyperref}
\usepackage[font=footnotesize]{caption}
\IEEEoverridecommandlockouts                              

\title{\LARGE \bf
EL-AGHF: Extended Lagrangian Affine Geometric Heat Flow*
}

\author{Sangmin Kim$^{1}$ and Hae-Won Park$^{1}$, \textit{Member, IEEE}
\thanks{This work was supported by the Technology Innovation Program(or Industrial Strategic Technology Development Program-Robot Industry Technology Development)(00427719, Dexterous and Agile Humanoid Robots for Industrial Applications) funded By the Ministry of  Trade Industry \& Energy(MOTIE, Korea)}
\thanks{$^{1}$All authors are with the Humanoid Robot Research Center, Korea
Advanced Institute of Science and Technology, Daejeon 34141, Korea.
        {\tt\small haewonpark@kaist.ac.kr}}%
}

\begin{document}

\maketitle
\thispagestyle{empty}
\pagestyle{empty}

\begin{abstract}

We propose a constrained Affine Geometric Heat Flow (AGHF) method that evolves so as to suppress the dynamics gaps associated with inadmissible control directions. AGHF provides a unified framework applicable to a wide range of motion planning problems, including both holonomic and non-holonomic systems. However, to generate admissible trajectories, it requires assigning infinite penalties to inadmissible control directions. This design choice, while theoretically valid, often leads to high computational cost or numerical instability when the penalty becomes excessively large. To overcome this limitation, we extend AGHF in an Augmented Lagrangian method approach by introducing a dual trajectory related to dynamics gaps in inadmissible control directions. This method solves the constrained variational problem as an extended parabolic partial differential equation defined over both the state and dual trajectorys, ensuring the admissibility of the resulting trajectory. We demonstrate the effectiveness of our algorithm through simulation examples.

\end{abstract}

\section{INTRODUCTION}
Despite extensive efforts, motion planning for nonlinear and underactuated systems remains a fundamental challenge due to their inherent complexity. A general approach involves encoding system constraints into a metric, so that the resulting shortest trajectorys on the manifold naturally adhere to the system dynamics or achieve desired behaviors \cite{liu2019affine, belabbas2017new, fan2020geometric, adu2025bring, adu2025phasing, van2022geometric, van2024geometric, xie2020geometric, ratliff2018riemannian}.

The Affine Geometric Heat Flow (AGHF) \cite{liu2019affine} framework encodes dynamic constraints into a metric, formulating the motion planning problem as a curve-shortening variational problem. This variational problem is then transformed into a partial differential equation (PDE), which deforms any arbitrary trajectory between the initial and final states into a dynamically feasible trajectory that minimizes control effort.

Although AGHF has been successfully applied to various motion planning problems in simulation~\cite{adu2025bring, fan2020mid, liu2020geometric} and in real-world settings~\cite{adu2025phasing}, existing approaches still struggle to produce fully dynamically feasible solutions for underactuated systems. To eliminate violations of dynamics in inadmissible control directions—that is, directions in which actuation is not available—AGHF requires infinite metric scaling along those directions. However, such scaling is numerically intractable, and excessively large values often lead to computational instability and numerical issues.

To address the limitations of AGHF, Chen et al. \cite{chen2024motion} proposed an optimization-based framework that refines AGHF-generated trajectories to achieve dynamic feasibility. While this approach yields trajectories that are both more dynamically feasible and faster to compute than those from pure AGHF with high metric scaling, it can struggle to fine-tune the solution when the initial AGHF trajectory exhibits severe violations of system dynamics.

In this paper, we propose an Extended Lagrangian AGHF (EL-AGHF) method that aims to achieve dynamic feasibility in inadmissible control directions. By interpreting the dynamics in these directions as constraints, we formulate the motion planning problem as a constrained variational problem. We introduce dual trajectorys associated with the constraints and reformulate the problem as a min-max optimization of the extended Lagrangian's action functional. The variational problem is subsequently transformed into an extended parabolic PDE defined over both the state and dual trajectorys, which is progressively solved by a PDE solver to generate an admissible trajectory.


\section{Preliminaries} \label{sec:preliminaries}
This section presents the motion planning problem addressed in this study and provides an overview of AGHF, a variational approach employed to solve it.

\subsection{Affine Geometric Heat Flow}
Consider an affine system that is controllable:
\begin{align}
\dot{\mathbf{x}}(t) = F_d(\mathbf{x}(t)) + F(\mathbf{x}(t)) \mathbf{u}(t), \label{eq:dynamics}
\end{align}
where $\mathbf{x}(t) \in \mathbb{R}^n$ and $\mathbf{u}(t) \in \mathbb{R}^m$ denote the state and control input. $F_d(\mathbf{x}) \in \mathbb{R}^n$ is the drift term, and the columns of $F(\mathbf{x}) \in \mathbb{R}^{n \times m}$ represent the admissible control directions. 
To complement this, \( F_c(\mathbf{x}) \in \mathbb{R}^{n \times (n - m)} \) denotes the matrix of inadmissible control directions, whose columns, obtained for example via the Gram–Schmidt process, span the orthogonal complement to the column space of \( F(\mathbf{x}) \). Then, $
\bar{F}(\mathbf{x}) := \left[ F_c(\mathbf{x}) \;\middle|\; F(\mathbf{x}) \right] \in \mathbb{R}^{n \times n}
$ is constructed, and it is assumed to be full rank.

Given an initial state $\mathbf{x}_0$ and a final state $\mathbf{x}_f$ that lies within the reachable set of the system, our objective is to find a trajectory over the interval $[0, T]$ that satisfies the system dynamics~\eqref{eq:dynamics} while minimizing the control effort. This leads to the following variational problem:
\begin{align}
\min_{\mathbf{u}} \quad & \int_0^T \| \mathbf{u}(t) \|_2^2 \, dt \label{eq:ocp} \\
\text{subject to} \quad & \dot{\mathbf{x}}(t) = F_d(\mathbf{x}(t)) + F(\mathbf{x}(t)) \mathbf{u}(t), \label{eq:ocp_const} \\
& \mathbf{x}(0) = \mathbf{x}_0, \quad \mathbf{x}(T) = \mathbf{x}_f, \label{eq:ocp_endpoint_const}
\end{align}
where~\eqref{eq:ocp_const} holds for all \( t \in [0, T] \).

 By encoding the system dynamics and objective functional into a Riemannian metric, the problem can be equivalently viewed as a drift-modified curve-shortening problem on a Riemannian manifold, where the objective is to minimize the following action functional $\mathcal{A}(\mathbf{x}(\cdot))$ defined in terms of the Lagrangian $\mathcal{L}(\mathbf{x}(t), \dot{\mathbf{x}}(t))$:
\begin{align}
\mathcal{A}(\mathbf{x}(\cdot)) &= \int_0^T \mathcal{L}(\mathbf{x}(t), \dot{\mathbf{x}}(t)) \, dt, \label{eq:action}\\
\mathcal{L}(\mathbf{x}(t), \dot{\mathbf{x}}(t)) &= \left( \dot{\mathbf{x}}(t) - F_d(\mathbf{x}(t)) \right)^\top 
G(\mathbf{x}(t)) \notag \\
&\cdot \left( \dot{\mathbf{x}}(t) - F_d(\mathbf{x}(t)) \right), \label{eq:lagrangian} 
\end{align}
subject to the boundary conditions $\mathbf{x}(0) = \mathbf{x}_0, \; \mathbf{x}(T) = \mathbf{x}_f$.

Here, the Riemannian metric tensor \( G(\mathbf{x}(t)) \) is given by
\begin{align}
G(\mathbf{x}(t)) = \left( \bar{F}(\mathbf{x}(t))^{-1} \right)^\top D \, \bar{F}(\mathbf{x}(t))^{-1},
\end{align}
with the scaling matrix \( D = \mathrm{diag}(\lambda, \dots, \lambda, 1, \dots, 1) \in \mathbb{R}^{n \times n} \) for \(\lambda > 0\), where the first \( n - m \) entries correspond to inadmissible and the remaining \( m \) to admissible control directions.
 In the limit as \( \lambda \to \infty \), the minimizer trajectory of the action functional \( \mathcal{A}(\mathbf{x}(\cdot)) \) becomes the solution to the variational problem~\eqref{eq:ocp}–\eqref{eq:ocp_endpoint_const}.

The AGHF framework solves a parabolic PDE that progressively minimizes the action functional $\mathcal{A}(\mathbf{x}(\cdot,s))$ along an artificial evolution variable $s$. To this end, the state trajectory is extended to a two-variable function $\mathbf{x}(t, s)$, defined on the domain $[0, T] \times [0, \infty)$. As the PDE is integrated forward in the $s$-direction, the trajectory $\mathbf{x}(t, s)$, starting from an arbitrary non-feasible initial trajectory $\mathbf{x}(t, 0)$, is gradually deformed into a curve that minimizes $\mathcal{A}(\mathbf{x}(\cdot, s))$. The AGHF is given by
\begin{align}
\frac{\partial \mathbf{x}}{\partial s}(t, s) =& G^{-1}(\mathbf{x}(t, s)) 
\Big( 
\frac{d}{dt}\frac{\partial \mathcal{L}}{\partial \dot{\mathbf{x}}}(\mathbf{x}(t, s), \dot{\mathbf{x}}(t, s)) \notag \\
&- \frac{\partial \mathcal{L}}{\partial \mathbf{x}}(\mathbf{x}(t, s), \dot{\mathbf{x}}(t, s)) 
\Big),
\label{eq:aghf}
\end{align}
with boundary conditions $\mathbf{x}(0, s) = \mathbf{x}_0$ and $\mathbf{x}(T, s) = \mathbf{x}_f$ for all $s \ge 0$.

According to Lemma 1 in \cite{liu2019affine}, the AGHF equation~\eqref{eq:aghf} ensures that the action functional is non-increasing along $s$:
\begin{align}
\frac{d}{ds} \mathcal{A}(\mathbf{x}(\cdot, s)) \le 0. \label{eq:aghf_convergence}
\end{align}
The trajectory $\mathbf{x}(t, s)$ converges to a limiting trajectory $\mathbf{x}^*(t)$ that satisfies the Euler–Lagrange equation, because $\mathcal{A}(\mathbf{x}(\cdot, s))$ is bounded below and~\eqref{eq:aghf_convergence} holds with equality if and only if $\frac{\partial \mathbf{x}}{\partial s} = 0$:
\begin{align}
\frac{d}{dt} \frac{\partial \mathcal{L}}{\partial \dot{\mathbf{x}}}(\mathbf{x}^*(t), \dot{\mathbf{x}}^*(t)) 
- \frac{\partial \mathcal{L}}{\partial \mathbf{x}}(\mathbf{x}^*(t), \dot{\mathbf{x}}^*(t)) = 0,
\end{align}
which defines a necessary condition for a local minimizer of the action functional $\mathcal{A}(\mathbf{x}(\cdot))$.

The control input trajectory corresponding to a given trajectory \( \mathbf{x}^*(t) \) can be evaluated as
\begin{align}
\tilde{\mathbf{u}}(t) := F(\mathbf{x}^*(t))^{\dagger} \left( \dot{\mathbf{x}}^*(t) - F_d(\mathbf{x}^*(t)) \right),
\end{align}
where \( (\cdot)^{\dagger} \) denotes the pseudoinverse.
According to Theorem 1 in \cite{liu2019affine}, there exists a constant \( C > 0 \) such that for any \( \lambda > 0 \), the terminal error \(e(T)\) satisfies
\begin{align}
e(T)=\| \tilde{\mathbf{x}}(T) - \mathbf{x}_f \|_2 \le \sqrt{\frac{C}{\lambda}}, \label{terminal_error}
\end{align}
where \( \tilde{\mathbf{x}}(t) \) denotes the trajectory obtained by integrating the system dynamics~\eqref{eq:dynamics} under the control input \( \tilde{\mathbf{u}}(t) \). From~\eqref{terminal_error}, it follows that ensuring the solution path \( \mathbf{x}^*(t) \) is dynamically feasible requires taking the limit \( \lambda \to \infty \).

\section{Method}

In this section, we propose EL-AGHF, a novel methodology for addressing dynamics violations. We also extend our approach to handle additional kinematic constraints.

\subsection{Extended Lagrangian AGHF} \label{subsec:el_aghf}
Achieving dynamics feasibility of the trajectory $\mathbf{x}^*(t)$ obtained through AGHF may require increasing $\lambda$ to a large value, which can lead to solving PDEs that are computationally expensive and potentially numerically unstable~\cite{fan2020mid, chen2024motion}. To address this issue, we extend AGHF by introducing a dual trajectory \(\boldsymbol{\mu}(t) \in \mathbb{R}^{n-m}\) associated with the dynamics constraints in the inadmissible control directions to guarantee dynamics feasibility.

We reformulate the variational problem of minimizing~\eqref{eq:action} as a pointwise constrained variational problem, given by:
\begin{align}
\min_{\mathbf{x}} \quad & \mathcal{A}(\mathbf{x}(\cdot)) \label{eq:constrained_var} \\
\text{subject to} \quad & F_c(\mathbf{x}(t))^{\dagger} \left( \dot{\mathbf{x}}(t) - F_d(\mathbf{x}(t)) \right) = \mathbf{0}, \label{eq:constrained_var_const} \\
& \mathbf{x}(0) = \mathbf{x}_0, \quad \mathbf{x}(T) = \mathbf{x}_f, \label{eq:constrained_var_endpoint_const}
\end{align}
where~\eqref{eq:constrained_var_const} holds for all \( t \in [0, T] \).
The action~\eqref{eq:constrained_var} coincides with the original definition given by~\eqref{eq:action}--\eqref{eq:lagrangian}.
The solution trajectory to~\eqref{eq:constrained_var}--\eqref{eq:constrained_var_endpoint_const} satisfies~\eqref{eq:constrained_var_const},
which guarantees that the components of the Lagrangian~\eqref{eq:lagrangian} associated with the inadmissible control directions always vanish.
Therefore, excluding these components from the objective functional does not alter the solution.
However, to satisfy the strengthened Legendre condition~\cite{van2004second}, we retain these components as an augmented term:
\begin{align}
\frac{\partial^2 \mathcal{L}}{\partial \dot{\mathbf{x}} \, \partial \dot{\mathbf{x}}} \succ \mathbf{0}.
\label{eq:strict_legendre_condition}
\end{align}
The constrained variational problem~\eqref{eq:constrained_var}--\eqref{eq:constrained_var_endpoint_const} can be equivalently viewed as a min-max optimization problem of the action \(\bar{\mathcal{A}}(\mathbf{x}(\cdot), \boldsymbol{\mu}(\cdot))\) defined over the extended Lagrangian \(\bar{\mathcal{L}}(\mathbf{x}(t), \dot{\mathbf{x}}(t), \boldsymbol{\mu}(t))\) with respect to the state trajectory \(\mathbf{x}(t)\) and the dual trajectory \(\boldsymbol{\mu}(t)\)~\cite{van2004holonomic, liberzon2011calculus}:
\begin{align}
\bar{\mathcal{A}}(\mathbf{x}(\cdot), \boldsymbol{\mu}(\cdot)) 
&= \int_0^T \bar{\mathcal{L}}(\mathbf{x}(t), \dot{\mathbf{x}}(t), \boldsymbol{\mu}(t)) \, dt, \label{eq:action_extended} \\
\bar{\mathcal{L}}(\mathbf{x}, \dot{\mathbf{x}}, \boldsymbol{\mu}) 
&= \left( \dot{\mathbf{x}} - F_d \right)^\top G \left( \dot{\mathbf{x}} - F_d \right) \notag \\
&\quad + 2\lambda \boldsymbol{\mu}^\top F_c^{\dagger} \left( \dot{\mathbf{x}} - F_d \right) \notag \\
&= \left( \dot{\mathbf{x}} - F_d + F_c \boldsymbol{\mu} \right)^\top G \left( \dot{\mathbf{x}} - F_d + F_c \boldsymbol{\mu} \right) \notag \\
&\quad - \lambda \boldsymbol{\mu}^\top \boldsymbol{\mu}, \label{eq:lagrangian_extended}
\end{align}
where \( F_d = F_d(\mathbf{x}(t)) \), \( F_c = F_c(\mathbf{x}(t)) \), and \( G = G(\mathbf{x}(t)) \).

As in the AGHF method, we introduce an artificial evolution variable \(s\) and solve a parabolic PDE by integrating along the \(s\)-direction to progressively minimize and maximize the extended action \(\bar{\mathcal{A}}(\mathbf{x}(\cdot,s), \boldsymbol{\mu}(\cdot,s))\) with respect to \(\mathbf{x}(\cdot,s)\) and \(\boldsymbol{\mu}(\cdot,s)\), respectively.  
The resulting PDE system, referred to as our EL-AGHF, is given by:
\begin{align}
\frac{\partial \mathbf{x}}{\partial s}(t, s) =& G^{-1}(\mathbf{x}(t, s)) 
\Big( 
\frac{d}{dt}\frac{\partial \bar{\mathcal{L}}}{\partial \dot{\mathbf{x}}}(\mathbf{x}(t, s), \dot{\mathbf{x}}(t, s),\boldsymbol{\mu}(t,s)) \notag \\
&- \frac{\partial \bar{\mathcal{L}}}{\partial \mathbf{x}}(\mathbf{x}(t, s), \dot{\mathbf{x}}(t, s),\boldsymbol{\mu}(t,s)) 
\Big),
\label{eq:el-aghf-x}\\
\frac{\partial \boldsymbol{\mu}}{\partial s}(t, s) =& \left(F_c(\mathbf{x}(t, s))^\top G(\mathbf{x}(t, s)) F_c(\mathbf{x}(t, s))\right)^{-1} \notag\\
&\cdot \frac{\partial \bar{\mathcal{L}}}{\partial \boldsymbol{\mu}}(\mathbf{x}(t, s), \dot{\mathbf{x}}(t, s),\boldsymbol{\mu}(t,s)),
\label{eq:el-aghf-mu}
\end{align}
with boundary conditions $\mathbf{x}(0, s) = \mathbf{x}_0$ and $\mathbf{x}(T, s) = \mathbf{x}_f$ for all $s \ge 0$. 

Following the same argument of Lemma 1 in~\cite{liu2019affine}, when the dual trajectory $\boldsymbol{\mu}$ is fixed at $\boldsymbol{\mu}^{\star}$, the augmented action functional $\bar{\mathcal{A}}(\mathbf{x}(\cdot, s), \boldsymbol{\mu}^{\star})$ is non-increasing along the $s$-direction:
\begin{align}
\frac{d}{ds} \bar{\mathcal{A}}(\mathbf{x}(\cdot, s), \boldsymbol{\mu}^{\star}) \le 0. \label{eq:extended_aghf_convergence}
\end{align}
If the trajectories $\mathbf{x}(t, s)$ and $\boldsymbol{\mu}(t, s)$ converge to $\mathbf{x}^{*}(t)$ and $\boldsymbol{\mu}^{*}(t)$ at $s = s_{\max}$, then the limit trajectories satisfy the following stationarity conditions associated with the constrained variational problem~\eqref{eq:constrained_var}--\eqref{eq:constrained_var_endpoint_const}:
\begin{align}
\left( \frac{\partial}{\partial \mathbf{x}} - \frac{d}{dt} \frac{\partial}{\partial \dot{\mathbf{x}}} \right) \bar{\mathcal{L}}(\mathbf{x}^{*}(t), \dot{\mathbf{x}}^{*}(t), \boldsymbol{\mu}^{*}(t)) &= \mathbf{0}, \label{eq:stationarity_euler_lagrange} \\
F_c(\mathbf{x}^{*}(t))^{\dagger} \left( \dot{\mathbf{x}}^{*}(t) - F_d(\mathbf{x}^{*}(t)) \right) &= \mathbf{0}. \label{eq:stationarity_constraint}
\end{align}
The trajectory $\mathbf{x}^*(t)$ is a local minimum of~\eqref{eq:constrained_var}--\eqref{eq:constrained_var_endpoint_const} that ensures dynamic feasibility.

\subsection{EL-AGHF with Kinematic Constraints}

In AGHF, additional kinematic constraints can be incorporated either by modifying the Riemannian metric tensor via a barrier function applied to \( G(x(t)) \)~\cite{liu2019affine, liu2019homotopy}, or by augmenting the original Lagrangian~\eqref{eq:lagrangian} with penalty terms that penalize constraint violations~\cite{fan2020mid, adu2025bring, adu2025phasing}:
\begin{align}
\mathcal{L}^{\text{c}}(\mathbf{x}, \dot{\mathbf{x}}) = \mathcal{L}(\mathbf{x}, \dot{\mathbf{x}}) + \sum_{j \in \mathcal{J}} b(h_j(\mathbf{x})),
\end{align}
where \( \mathcal{J} \) denotes the index set of all kinematic constraints, and the penalty term is defined as
\begin{align}
b(h_j(\mathbf{x})) = \lambda^{\text{c}}  (h_j(\mathbf{x}))^2  S(h_j(\mathbf{x})),
\end{align}
with \( \lambda^{\text{c}} > 0 \) as the penalty coefficient, and the switching function \( S(h_j(\mathbf{x})) \) is given by
\begin{align}
S(h_j(\mathbf{x})) =
\begin{cases}
1, & \text{if } h_j(\mathbf{x}) = 0, \\
H(h_j(\mathbf{x})), & \text{if } h_j(\mathbf{x}) \leq 0,
\end{cases}
\end{align}
with \( H(h_j(\mathbf{x})) \) being a smooth approximation of the Heaviside step function satisfying \( H(z) \approx 0 \) if \( z < 0 \) and \( H(z) \approx 1 \) if \( z > 0 \).

The former, barrier function-based method can cause numerical instability due to large values in the metric tensor near constraint boundaries. The latter, penalty term-based method may fail to enforce constraints when the penalty coefficient \( \lambda^{\text{c}} \) is small, and can similarly lead to numerical issues when \( \lambda^{\text{c}} \) is too large.

We extend the method presented in Section~\ref{subsec:el_aghf} by introducing a dual trajectory \( \mu^c_j(t) \in \mathbb{R} \) for each kinematic constraint. The associated constraint terms are added to the extended Lagrangian~\eqref{eq:lagrangian_extended}:
\begin{align}
\bar{\mathcal{L}}^c&(\mathbf{x}, \dot{\mathbf{x}}, \boldsymbol{\mu}, \boldsymbol{\mu}^c) 
= \bar{\mathcal{L}}(\mathbf{x}, \dot{\mathbf{x}}, \boldsymbol{\mu}) \nonumber \\
&\quad + \sum_{j \in \mathcal{J}} \lambda^{\text{c}} \left(\left(h_j(\mathbf{x}) + \mu^c_j \right)^2-(\mu^c_j)^2\right) S(h_j(\mathbf{x})),
\label{eq:constrinaed_lagrangian_extended}
\end{align}
where \( \boldsymbol{\mu}^c := [\mu^c_1, \mu^c_2, \dots, \mu^c_{|\mathcal{J}|}]^\top \) is the stacked vector of dual variables for the kinematic constraints.

The resulting PDE system, referred to as our EL-AGHF with kinematic constraints, is given by:
\begin{align}
\frac{\partial \mathbf{x}}{\partial s}(t, s) 
=&\; G^{-1} \left( 
\frac{d}{dt}\frac{\partial \bar{\mathcal{L}}^c}{\partial \dot{\mathbf{x}}} 
- \frac{\partial \bar{\mathcal{L}}^c}{\partial \mathbf{x}} 
\right), \label{eq:el-aghf-x} \\[1ex]
\frac{\partial \boldsymbol{\mu}}{\partial s}(t, s) 
=&\; \left(F_c^\top G F_c\right)^{-1} 
\frac{\partial \bar{\mathcal{L}}^c}{\partial \boldsymbol{\mu}}, \label{eq:el-aghf-mu-const} \\[1ex]
\frac{\partial \mu^c_j}{\partial s}(t, s) 
=&\; \frac{1}{\lambda^{\text{c}}} 
\frac{\partial \bar{\mathcal{L}}^c}{\partial \mu^c_j}, 
\quad \text{for all } j \in \mathcal{J}. \label{eq:el-aghf-mu_c}
\end{align}

\section{Experiments and Results}

In this section, we compare the proposed EL-AGHF method with AGHF across various scenarios. First, we evaluate on a unicycle model with constant linear velocity and on a dynamic unicycle, both without kinematic constraints, as in~\cite{liu2019affine}. Then, we compare them in a mid-air motion planning setting with kinematic constraints, following~\cite{fan2020mid}.

\subsection{Experimental Setup}

We conducted our experiments based on the code used in~\cite{liu2019affine} and~\cite{fan2020mid}. The \texttt{pdepe} toolbox in MATLAB was used to solve the PDE. To numerically integrate the parabolic PDE, boundary conditions for equations~\eqref{eq:el-aghf-mu}, \eqref{eq:el-aghf-mu-const}, and~\eqref{eq:el-aghf-mu_c} are required; hence, free boundary conditions were applied. 

The solver terminated integration upon satisfying the following convergence criteria for all \( k \in \mathcal{K} \), where \( \mathcal{K} \) denotes the set of discretized time indices used for numerically solving the PDE, and \( \epsilon > 0 \) is a predefined convergence threshold:
\begin{align}
\left\| \frac{\partial \mathbf{x}}{\partial s}(t_k, s) \right\|_\infty &< \epsilon, \label{eq:state_convergence} \\
\left\| \frac{\partial \boldsymbol{\mu}}{\partial s}(t_k, s) \right\|_\infty &< \epsilon, \\
\left\| \frac{\partial \boldsymbol{\mu}^c}{\partial s}(t_k, s) \right\|_\infty &< \epsilon.
\end{align}

\subsection{Unicycle Model} \label{subsec:Unicycle Model}
The system under consideration is a unicycle moving on a planar surface without slipping. All unicycle model experiments used a fixed convergence threshold of \( \epsilon = 10^{-4} \).

1) Unicycle of constant linear velocity:
We consider a planar unicycle model with unit constant linear velocity. The system dynamics are given by:
\begin{align}
\underbrace{
\begin{bmatrix}
\dot{x} \\
\dot{y} \\
\dot{\theta}
\end{bmatrix}
}_{\dot{\mathbf{x}}}
=
\underbrace{
\begin{bmatrix}
\cos \theta \\
\sin \theta \\
0
\end{bmatrix}
}_{F_d}
+
\underbrace{
\begin{bmatrix}
0 \\
0 \\
1
\end{bmatrix}
}_{F}
u, \label{eq:Unicycle_of_constant_linear_velocity}
\end{align}
where \( x \) and \( y \) denote the position along the \( x \)- and \( y \)-axes, respectively, and \( \theta \) represents the orientation of the car body with respect to the inertial frame.  
Here, \( \bar{F}(\mathbf{x}) = \left[ F_c(\mathbf{x}) \;\middle|\; F(\mathbf{x}) \right] \in \mathbb{R}^{3 \times 3} \) is chosen to be the identity matrix.

The boundary conditions are set to \( \mathbf{x}_0 = [0,\ 0,\ 0]^\top \) and \( \mathbf{x}_f = [0,\ 1,\ 0]^\top \), with a total time duration \( T = 5 \).  
Both EL-AGHF and AGHF use a linearly interpolated initial state trajectory defined as  
\( \mathbf{x}(t,0) = \mathbf{x}_0 + (\mathbf{x}_f - \mathbf{x}_0)t/T \), and EL-AGHF initializes the dual trajectory as \( \boldsymbol{\mu}(t,0) = \mathbf{0} \).

\begin{figure}[t!]
      \centering
      \includegraphics[width=0.47\textwidth]{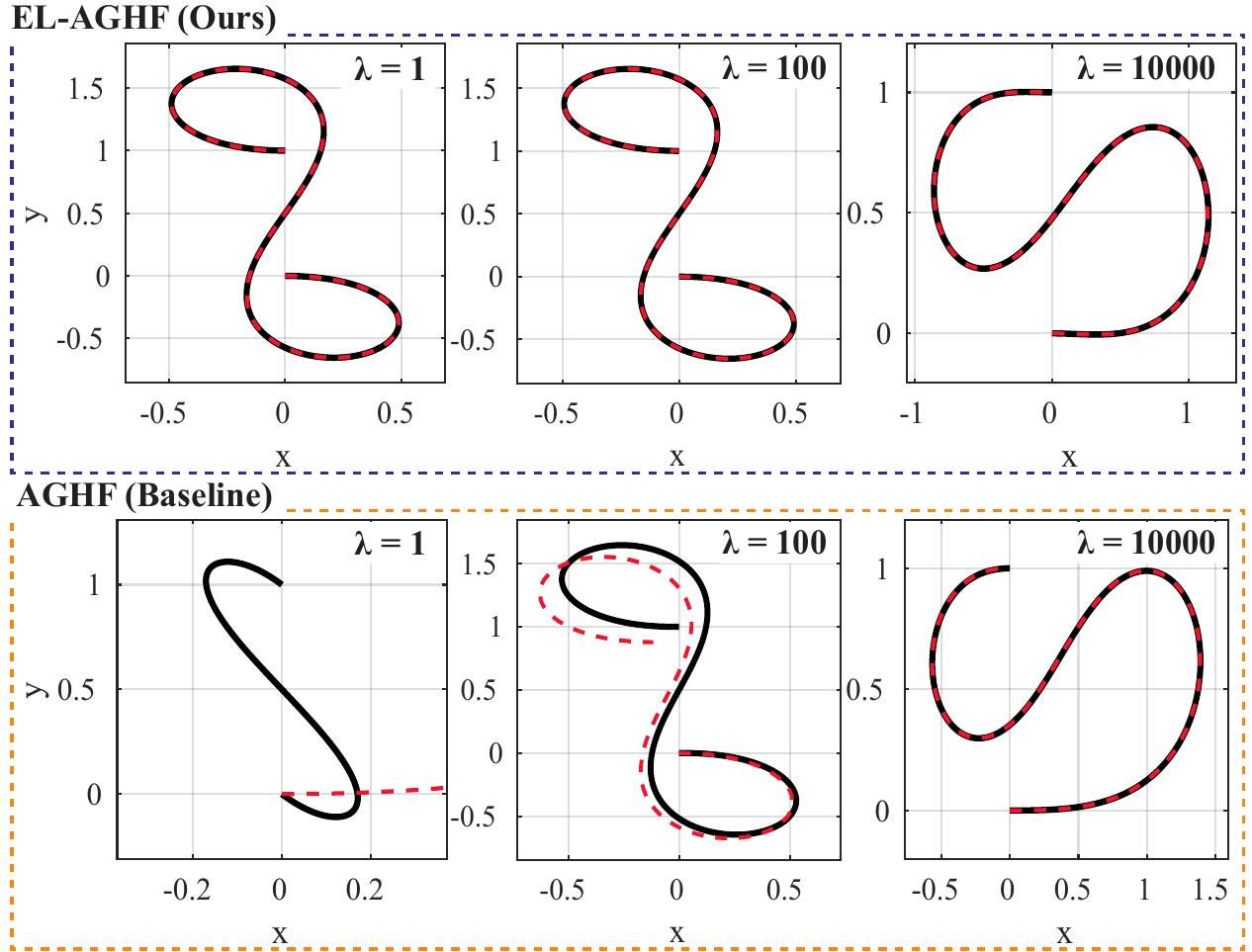}
      \caption{Trajectories generated by EL-AGHF and AGHF for the unicycle model with constant linear velocity~\eqref{eq:Unicycle_of_constant_linear_velocity}. The black dashed line represents the solution state trajectory, while the red dotted line denotes the trajectory obtained by integrating the system dynamics under the corresponding control input.}
    \label{fig:Unicycle_constant_vel}
\end{figure}

The results generated by EL-AGHF and AGHF for different values of \( \lambda \) are presented in Figure~\ref{fig:Unicycle_constant_vel} and Table~\ref{tab:const_vel}.  
EL-AGHF consistently maintains small dynamics gaps across all tested values of \( \lambda \).
In contrast, AGHF exhibits large dynamics gaps for small \( \lambda \) values. Although increasing \( \lambda \) reduces the dynamics gap, it also leads to numerical instability, resulting in larger values of \( s_{\max} \) and increased solving time.

\begin{table}[]
\caption{Comparison on the unicycle model with constant linear velocity.
}
\centering
\resizebox{0.98\linewidth}{!}{%
\begin{tabular}{|ll|l|l|l|l|l|}
\hline
\multicolumn{2}{|l|}{\(\lambda\)}                                      & 1      & 10    & 100    & 1000   & 10000   \\ \hline
\multicolumn{1}{|l|}{\multirow{3}{*}{EL-AGHF (Ours)}}  & \(s_{\max}\)     & 184.1 & 69.9 & 290.3 & 625.6 & 2787.2 \\ \cline{2-7} 
\multicolumn{1}{|l|}{}                                 & Time [s] & 0.28   & 0.18  & 1.26   & 1.26   & 3.13    \\ \cline{2-7} 
\multicolumn{1}{|l|}{}                                 & \(e(T)\)     & 5e-4   & 4e-4  & 3e-4   & 3e-4   & 3e-4    \\ \hline
\multicolumn{1}{|l|}{\multirow{3}{*}{AGHF (Baseline)}} & \(s_{\max}\)     & 27.4  & 24.8 & 366.4 & 952.3 & 6221.8 \\ \cline{2-7} 
\multicolumn{1}{|l|}{}                                 & Time [s] & 0.10   & 0.13  & 0.76   & 0.84   & 2.93    \\ \cline{2-7} 
\multicolumn{1}{|l|}{}                                 & \(e(T)\)     & 4.31   & 1.14  & 0.17   & 0.02   & 2e-3    \\ \hline
\end{tabular}
}
\label{tab:const_vel}
\end{table}

2) Dynamic unicycle:
We consider a planar unicycle with second-order dynamics. The system dynamics are given by:

\begin{align}
\underbrace{
\begin{bmatrix}
\dot{x} \\
\dot{y} \\
\dot{\theta} \\
\dot{v}_1 \\
\dot{v}_2
\end{bmatrix}
}_{\dot{\mathbf{x}}}
=
\underbrace{
\begin{bmatrix}
v_1 \cos \theta \\
v_1 \sin \theta \\
v_2 \\
0 \\
0
\end{bmatrix}
}_{F_d}
+
\underbrace{
\begin{bmatrix}
0 & 0 \\
0 & 0 \\
0 & 0 \\
1 & 0 \\
0 & 1
\end{bmatrix}
}_{F}
\underbrace{
\begin{bmatrix}
u_1 \\
u_2
\end{bmatrix}
}_{\mathbf{u}}. \label{eq:Dynamic_Unicycle}
\end{align}
where \( v_1 \) and \( v_2 \) denote the linear and angular velocities of the vehicle, respectively.
Here, \( \bar{F}(\mathbf{x}) = \left[ F_c(\mathbf{x}) \;\middle|\; F(\mathbf{x}) \right] \in \mathbb{R}^{5 \times 5} \) is chosen to be the identity matrix.

The boundary conditions are \( \mathbf{x}_0 = [0,\ 0,\ 0,\ 0,\ 0]^\top \) and \( \mathbf{x}_f = [0,\ 1,\ 0,\ 0,\ 0]^\top \), with duration \( T = 10 \).  
Both EL-AGHF and AGHF use a linearly interpolated trajectory between the boundary conditions (except for \( x(t,0) \)) as the initial trajectory.
The \( x \)-component follows a sinusoidal trajectory \( x(t,0) = 0.0001 \sin(\pi t / T) \), and EL-AGHF initializes the dual trajectory to zero, as in Section~\ref{subsec:Unicycle Model}-1).

\begin{figure}[t!]
      \centering
      \includegraphics[width=0.47\textwidth]{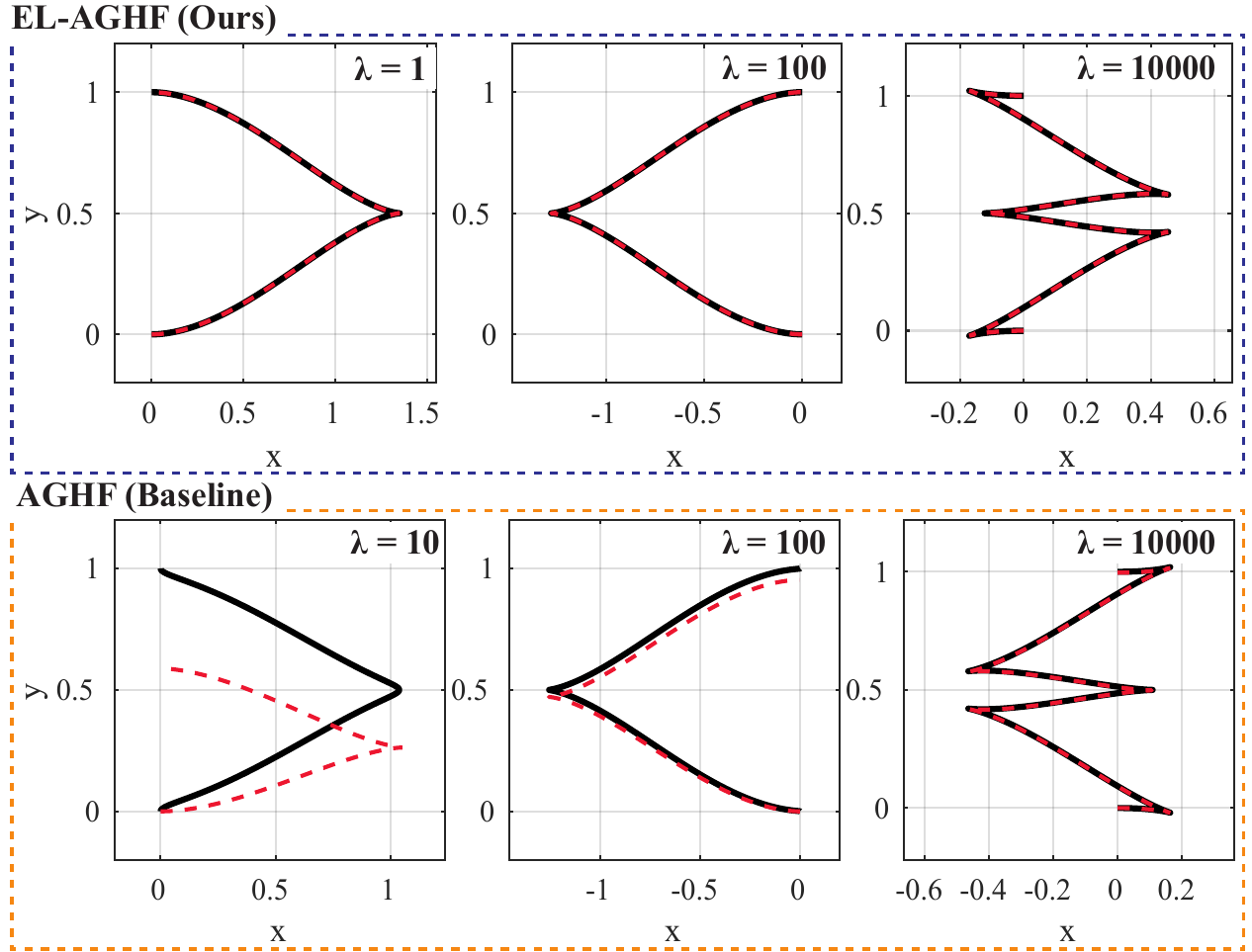}
      \caption{Trajectories generated by EL-AGHF and AGHF trajectories for the dynamic unicycle model~\eqref{eq:Dynamic_Unicycle}.}
    \label{fig:Unicycle_dynamic}
\end{figure}

The results for different values of \( \lambda \) are presented in Figure~\ref{fig:Unicycle_dynamic} and Table~\ref{tab:dynamic}.  
A similar trend to Section~\ref{subsec:Unicycle Model}-1) is observed, and EL-AGHF consistently achieves sufficiently small dynamics gaps across all values of \( \lambda \).  
For AGHF, an additional convergence condition \( s > 500 \) was introduced to avoid premature termination, as \eqref{eq:state_convergence} was locally satisfied during the early stage of \( s \).  
The case \( \lambda = 1 \) was excluded from the results because \eqref{eq:state_convergence} was satisfied throughout the entire range \( s < 10000 \), and the trajectory showed negligible evolution with respect to \( s \) compared to the initial trajectory.

\begin{table}[]
\caption{Comparison of results on the dynamic unicycle model.}
\centering
\resizebox{0.98\linewidth}{!}{%
\begin{tabular}{|ll|l|l|l|l|l|}
\hline
\multicolumn{2}{|l|}{\(\lambda\)}                                      & 1      & 10    & 100    & 1000   & 10000   \\ \hline
\multicolumn{1}{|l|}{\multirow{3}{*}{EL-AGHF (Ours)}}  & \(s_{\max}\)     & 116.2 & 355.3 & 1816.8 & 5837.3 & 2639.3 \\ \cline{2-7} 
\multicolumn{1}{|l|}{}                                 & Time [s] & 0.48   & 0.63  & 0.88   & 1.86   & 2.07    \\ \cline{2-7} 
\multicolumn{1}{|l|}{}                                 & \(e(T)\)     & 1e-4   & 2e-4  & 2e-4   & 2e-4   & 8e-4    \\ \hline
\multicolumn{1}{|l|}{\multirow{3}{*}{AGHF (Baseline)}} & \(s_{\max}\)     & \multicolumn{1}{c|}{--}  & 744.1 & 1985.6 & 5291.4 & 2890.8 \\ \cline{2-7} 
\multicolumn{1}{|l|}{}                                 & Time [s] & \multicolumn{1}{c|}{--}   & 0.12  & 0.26   & 0.73   & 0.73    \\ \cline{2-7} 
\multicolumn{1}{|l|}{}                                 & \(e(T)\)     & \multicolumn{1}{c|}{--}   & 0.42  & 0.05   & 7e-3   & 4e-3    \\ \hline
\end{tabular}
}
\label{tab:dynamic}
\end{table}

\begin{figure}[t!]
      \centering
      \includegraphics[width=0.47\textwidth]{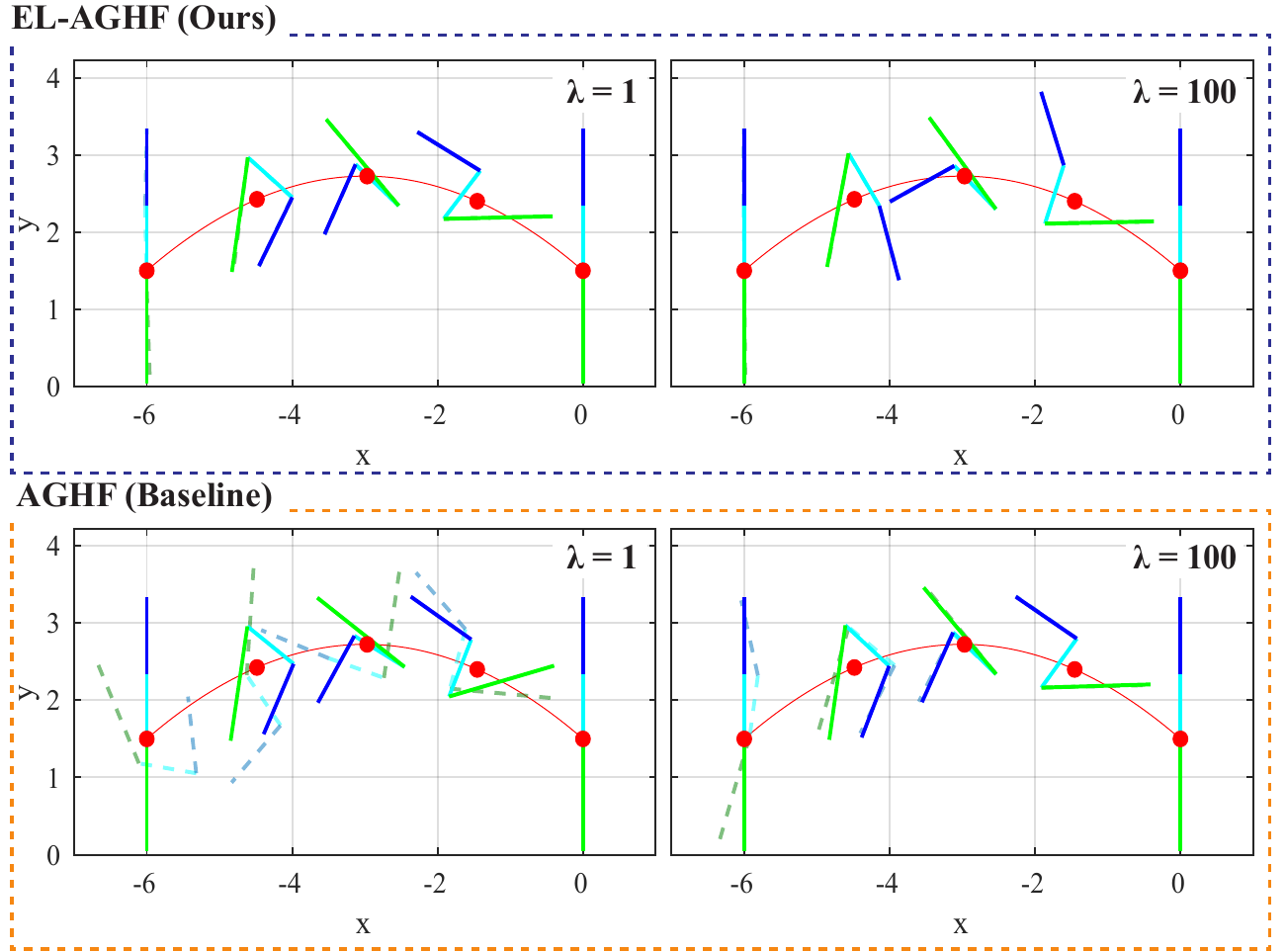}
      \caption{Snapshots at \( T = 0,\ 0.25,\ 0.5,\ 0.75,\ 1.0 \) of trajectories generated by EL-AGHF and AGHF for the unconstrained mid-air motion model. The dashed line represents the snapshot of the solution trajectory, while the dotted line shows the trajectory snapshot obtained by integrating the system dynamics under the corresponding control input. The red curves indicate the trajectories of the center of mass (COM), and the red dots denote the COM positions at the snapshot times. The COM motion is fully determined by the initial linear momentum, and the dynamics model~\eqref{eq:mid-air-motion-model} does not explicitly include the COM position.}
    \label{fig:Mid-Air_Motion_not_constrained}
\end{figure}
\begin{table}[]
\caption{Comparison on the unconstrained mid-air motion model.}
\centering
\resizebox{0.98\linewidth}{!}{%
\begin{tabular}{|ll|l|l|l|l|}
\hline
\multicolumn{2}{|l|}{\(\lambda\)}                                      & 1      & 10    & 100    & 1000   \\ \hline
\multicolumn{1}{|l|}{\multirow{2}{*}{EL-AGHF (Ours)}}  & Time [s]     & 12.8 & 52.2 & 600.0 & 600.0 \\ \cline{2-6} 
\multicolumn{1}{|l|}{}                                 & \(\hat{e}(T)\)     & 0.25   & 0.27  & 0.10   & 0.10    \\ \hline
\multicolumn{1}{|l|}{\multirow{2}{*}{AGHF (Baseline)}} & Time [s]     & 5.8  & 21.3 & 600.0 & 600.0 \\ \cline{2-6} 
\multicolumn{1}{|l|}{}                                 & \(\hat{e}(T)\)     & 16.40   & 10.89  & 1.02   & 0.26     \\ \hline
\end{tabular}
}
\label{tab:unconstrained}
\end{table}

\subsection{Mid-Air Motion Model}
\label{subsec:Mid-Air Model}
The system under consideration is a planar diver robot consisting of three links connected by revolute joints.  
The system dynamics are given by:
\begin{align}
\underbrace{
\begin{bmatrix}
\dot{\mathbf{q}} \\
\ddot{\mathbf{q}}
\end{bmatrix}
}_{\dot{\mathbf{x}}}
=
\underbrace{
\begin{bmatrix}
\dot{\mathbf{q}} \\
- D^{-1}\!(\mathbf{q})\, C(\mathbf{q}, \dot{\mathbf{q}})\dot{\mathbf{q}}
\end{bmatrix}
}_{F_d}
+
\underbrace{
\begin{bmatrix}
\mathbf{0}_{3\times2} \\
D^{-1}\!(\mathbf{q})\, \mathbf{E}
\end{bmatrix}
}_{F}
\underbrace{
\begin{bmatrix}
u_1 \\
u_2
\end{bmatrix}
}_{\mathbf{u}}, \label{eq:mid-air-motion-model}
\end{align}

where \( \mathbf{q} = [\theta_0,\ q_1,\ q_2]^\top \) denotes the reduced generalized coordinates, where \( \theta_0 \) represents the orientation of the base link in the ground inertial frame, and \( q_i \) denotes the \( i \)th joint angle.  
\( D(\mathbf{q}) \in \mathbb{R}^{3 \times 3} \) and \( C(\mathbf{q}, \dot{\mathbf{q}}) \in \mathbb{R}^{3 \times 3} \) are the inertia matrix and the Coriolis matrix defined with respect to the reduced coordinates.  
\( \mathbf{0}_{m \times n} \) denotes an \( m \times n \) zero matrix, and \( \mathbf{E} \in \mathbb{R}^{3 \times 2} \) denotes the selection matrix consisting of the last two columns of the \( 3 \times 3 \) identity matrix.  
Detailed definitions of \( D \) and \( C \) can be found in~\cite{fan2020mid}. Here, \( \bar{F}(\mathbf{x}) = \left[ F_c(\mathbf{x}) \;\middle|\; F(\mathbf{x}) \right] \in \mathbb{R}^{6 \times 6} \) is chosen as follows:
\[
F(\mathbf{x}) =
\begin{bmatrix}
I_3 & \mathbf{0}_{3 \times 3} \\
\mathbf{0}_{3 \times 3} & D^{-1}(\mathbf{q})
\end{bmatrix},
\]
where \( I_n \in \mathbb{R}^{n \times n} \) denotes the \( n \times n \) identity matrix.

The Mid-Air Motion Model is a system in which angular momentum is conserved, and thus a solution may not exist for arbitrary initial and final conditions~\cite{fan2020mid}. Therefore, the boundary conditions are partially fixed as \( \mathbf{x}_0 = [0,\ 0,\ 0,\ 0,\ 0,\ \cdot]^\top \) and \( \mathbf{x}_f = [2\pi,\ 0,\ 0,\ 0,\ \cdot,\ 0]^\top \), with a total duration of \( T = 1 \). Free boundary conditions are applied to the state components denoted by \( \cdot \). Furthermore, to measure the terminal error, we used \( \hat{e}(T) = \left\| \tilde{\mathbf{x}}(T) - \mathbf{x}(T) \right\|_2 \) instead of \( e(T) \), as defined in~\eqref{terminal_error}.

Both EL-AGHF and AGHF initialize the trajectory by linearly interpolating between the boundary conditions for \( \theta_0(t,0) \), while all other state components are set to zero. In EL-AGHF, the dual trajectory \( \boldsymbol{\mu}(t,0) \) is initialized to \( \mathbf{0} \). In Section~\ref{subsec:Mid-Air Model}-2), the constraint-associated dual trajectory \( \boldsymbol{\mu}^c(t,0) \) is also initialized to \( \mathbf{0} \). We used a convergence threshold of \( \epsilon = 10^{-2} \) in Section~\ref{subsec:Mid-Air Model}-1) and \( \epsilon = 10^{-1} \) in Section~\ref{subsec:Mid-Air Model}-2). Additionally, PDE integration was also terminated if the total solving time exceeded 600 seconds.

1) Unconstrained mid-air motion: First, we evaluate our method without additional kinematic constraints. Figure~\ref{fig:Mid-Air_Motion_not_constrained} and Table~\ref{tab:unconstrained} show the results of EL-AGHF and AGHF for different values of \( \lambda \). EL-AGHF maintained small dynamics violations across all values of \( \lambda \), and produced control inputs that enabled successful landing. However, AGHF failed to satisfy the system dynamics for low values of \( \lambda \); although the dynamics gap decreased as \( \lambda \) increased, both methods exhibited convergence issues in the high-\( \lambda \) regime and failed to satisfy the termination criteria within the time limit.


\begin{figure}[t!]
      \centering
      \includegraphics[width=0.47\textwidth]{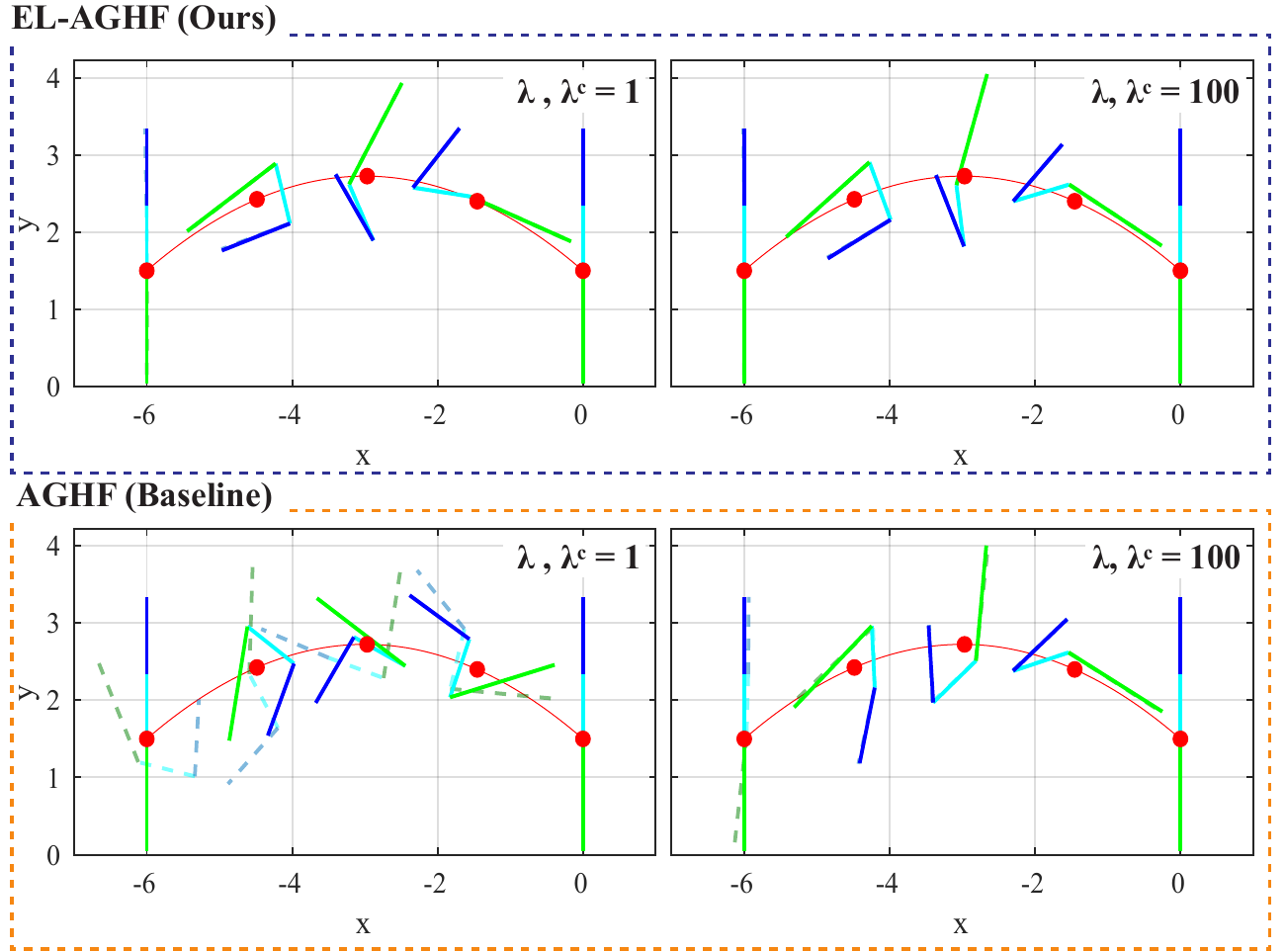}
      \caption{Snapshots at \( T = 0,\ 0.25,\ 0.5,\ 0.75,\ 1.0 \) of trajectories generated by EL-AGHF and AGHF for the constrained mid-air motion model.}
    \label{fig:Mid-Air_Motion_constrained}
\end{figure}

\begin{table}[]
\caption{Comparison on the constrained mid-air motion model.}
\centering
\resizebox{0.98\linewidth}{!}{%
\begin{tabular}{|ll|l|l|l|l|}
\hline
\multicolumn{2}{|l|}{\(\lambda,\ \lambda^c\)}                                      & 1      & 10    & 100    & 1000   \\ \hline
\multicolumn{1}{|l|}{\multirow{3}{*}{EL-AGHF (Ours)}}  & Time [s]     & 46.8 & 95.1 & 600.0 & 600.0 \\ \cline{2-6} 
\multicolumn{1}{|l|}{}                                 & \(\hat{e}(T)\) & 0.47   & 0.77  & 0.20   & 0.14      \\ \cline{2-6} 
\multicolumn{1}{|l|}{}                                 & \(e_{\text{viol}}\)     & 2e-3   & 2e-4  & 5e-5   & 7e-4    \\ \hline
\multicolumn{1}{|l|}{\multirow{3}{*}{AGHF (Baseline)}} & Time [s]    & 3.8  & 18.5 & 600.0 & 600.0 \\ \cline{2-6} 
\multicolumn{1}{|l|}{}                                 & \(\hat{e}(T)\) & 16.28   & 9.88  & 1.04   & 0.06      \\ \cline{2-6} 
\multicolumn{1}{|l|}{}                                 & \(e_{\text{viol}}\)     & 0.43   & 0.42  & 0.05   & 6e-3     \\ \hline
\end{tabular}
}
\label{tab:constrained}
\end{table}

2) Constrained mid-air motion: We impose the following inequality constraints on the system in~\eqref{eq:mid-air-motion-model}:
\begin{align}
h_1(\mathbf{x}) &= -q_2 + q_{\text{min}} \leq 0, \\
h_2(\mathbf{x}) &= q_2 - q_{\text{max}} \leq 0,
\end{align}
where a smooth approximation of the Heaviside step function is given by \( H(x) \approx \frac{1}{1 + e^{-k_s x}} \), and the constant \( k_s \) controls the sharpness of the approximation. We used \( q_{\text{min}} = -1.9 \), \( q_{\text{max}} = 1.9 \), and \( k_s = 100 \) in all experiments.

The results of both methods for different values of \( \lambda \) and \( \lambda^c \) are presented in Figure~\ref{fig:Mid-Air_Motion_constrained} and Table~\ref{tab:constrained}. For all experiments, \( \lambda \) and \( \lambda^c \) were assigned the same constant value. Similar to the results in Section~\ref{subsec:Mid-Air Model}-1), EL-AGHF exhibited smaller dynamics gaps than AGHF for low \( \lambda \), while both methods suffered from convergence issues at high penalty coefficients.
To evaluate constraint violations, we introduce the metric \( e_{\text{viol}} \), defined as
\begin{equation}
e_{\text{viol}} = \int_0^T \sum_{j=1}^2 \min\big(h_j(\mathbf{x}(t)),\, 0\big) \, dt.
\end{equation}
The metric shows that EL-AGHF also satisfied the kinematic constraints more effectively than AGHF.

\section{CONCLUSIONS}
In this work, we presented EL-AGHF, a constrained Affine Geometric Heat Flow method that extends the AGHF framework to address dynamic infeasibility. By interpreting the dynamics in inadmissible control directions as hard constraints and introducing associated dual trajectories, EL-AGHF reformulates the problem as a constrained variational formulation. The resulting extended parabolic PDE jointly evolves the state and the dual trajectory to progressively eliminate dynamics violations while maintaining a geometrically consistent formulation. We also applied the same approach to handle kinematic constraints.

Compared to standard AGHF, EL-AGHF avoids excessive metric scaling and improves numerical stability. Simulation results show that EL-AGHF produces dynamically admissible trajectories with smaller dynamics gaps and better constraint satisfaction, especially under low penalty coefficients. 

In future work, we plan to establish the convergence of the proposed method. Since our framework can be viewed as an extension of the Basic Differential Multiplier Method (BDMM) to constrained variational problems, we expect that its convergence can be analyzed through energy-based arguments, as in~\cite{platt1987constrained}. We also plan to improve the convergence rate of our method by introducing an automatic time-varying adjustment scheme for the penalty coefficients.

\section*{ACKNOWLEDGMENT}

We thank Professor S.~Liu~\cite{liu2019affine} and Dr.~Y.~Fan~\cite{fan2020mid} for generously providing their code and permitting its use.

\renewcommand{\bibfont}{\footnotesize}
\printbibliography

\end{document}